\documentclass[conference]{IEEEtran}
\IEEEoverridecommandlockouts

\usepackage{amsmath,amssymb,amsfonts}
\usepackage{algorithmic}
\usepackage{graphicx}
\usepackage{textcomp}
\usepackage{multirow} 
\usepackage{xcolor}
\usepackage{bm}
\usepackage{color}
 \usepackage{booktabs} 
\usepackage{flushend}
\usepackage{url}
\usepackage{cite}         
\usepackage{xcolor}
\usepackage[hidelinks, colorlinks=true, allcolors=blue]{hyperref}

\begin{document}

\title{SceneMixer: Exploring Convolutional Mixing Networks for Remote Sensing Scene Classification}

\author{
\centering
\IEEEauthorblockN{Mohammed~Q. Alkhatib}
\IEEEauthorblockA{College of Engineering and IT \\
University of Dubai\\
Emirates Road - Exit 49\\
Dubai 14143, UAE \\
email: mqalkhatib@ieee.org}

\and
\IEEEauthorblockN{Ali Jamali}
\IEEEauthorblockA{Department of Geography \\
Simon Fraser University\\
8888 University Dr W, Burnaby\\
BC V5A 1S6, Canada \\
email: alij@sfu.ca}
\and
\IEEEauthorblockN{Swalpa Kumar Roy}
\IEEEauthorblockA{Department of Computer \\
Science and Engineering \\
Tezpur University\\
Assam 784028, India \\
email: swalpa@tezu.ernet.in}
}
\maketitle

\begin{abstract}
Remote sensing scene classification plays a key role in Earth observation by enabling the automatic identification of land use and land cover (LULC) patterns from aerial and satellite imagery. Despite recent progress with convolutional neural networks (CNNs) and vision transformers (ViTs), the task remains challenging due to variations in spatial resolution, viewpoint, orientation, and background conditions, which often reduce the generalization ability of existing models. To address these challenges, this paper proposes a lightweight architecture based on the convolutional mixer paradigm. The model alternates between spatial mixing through depthwise convolutions at multiple scales and channel mixing through pointwise operations, enabling efficient extraction of both local and contextual information while keeping the number of parameters and computations low. Extensive experiments were conducted on the AID and EuroSAT benchmarks. The proposed model achieved overall accuracy, average accuracy, and Kappa values of 74.7\%, 74.57\%, and 73.79 on the AID dataset, and 93.90\%, 93.93\%, and 93.22 on EuroSAT, respectively. These results demonstrate that the proposed approach provides a good balance between accuracy and efficiency compared with widely used CNN- and transformer-based models.
Code will be publicly available on: \textcolor{blue}{https://github.com/mqalkhatib/SceneMixer}
\end{abstract}
\begin{IEEEkeywords}
Remote Sensing Scene Classification, Convolutional Mixer, Lightweight Deep Learning, Depth-Wise Convolution, Aerial Imagery.

\end{IEEEkeywords}

\section{Introduction}
\label{sec:intro}
Remote sensing scene classification is a fundamental task in Earth observation, aiming to automatically assign semantic labels to aerial or satellite images based on land use and land cover (LULC) patterns \cite{hong2023spectralgpt}. Unlike pixel-level classification, which focuses on individual spectral signatures \cite{alkhatib2024hsiformer}, scene classification utilizes the overall spatial, spectral, and structural characteristics of image patches. This makes it valuable for applications such as urban planning \cite{nielsen2015classification}, and agriculture \cite{zhang2020applications}. Deep learning has significantly advanced this field, with convolutional neural networks (CNNs) and vision transformers (ViTs) achieving strong performance on complex benchmarks \cite{wu2021convolutional}. Nevertheless, classification remains challenging due to variations in spatial resolution, viewpoint, orientation, translation, and background complexity \cite{zhao2020remote}.  

CNNs excel at local feature extraction through equivariant representations, sparse interaction, and weight sharing, enabling robust recognition even under shifts or rotations \cite{wang2022transferring}. However, their limited receptive fields hinder global context modeling \cite{jamali2023wetmapformer}. ViTs address this by capturing long-range dependencies but suffer from high computational cost and lack of inductive bias, which limit their effectiveness in small-scale or resource-constrained scenarios \cite{jamali2025mshcct}. To overcome these issues, hybrid models have been proposed. The Multiscale Multihead Compact Convolutional Transformer (MSHCCT) enhances multiscale representation using convolutional tokenization with lightweight transformers \cite{jamali2025mshcct}, while CTNet jointly exploits CNN and ViT streams to combine local structural and semantic features, achieving strong results on AID and NWPU-RESISC45 \cite{deng2021cnns}. ViCxLSTM further integrates CNNs, ViTs, and extended long short-term memory (xLSTM) modules to simultaneously capture local textures, long-range dependencies, and global context, establishing new benchmarks in aerial scene classification \cite{roy2025vicxlstm}.  

\begin{figure*} [t!]
   \centering
   \includegraphics[clip=true, trim = 20 10 20 5,width= 0.9\linewidth]{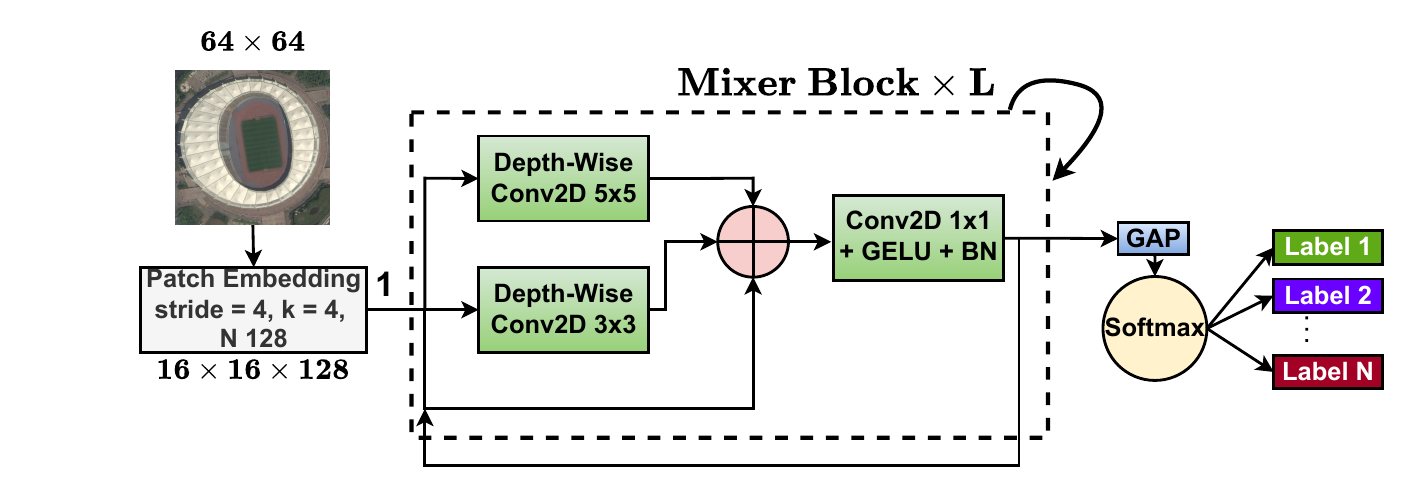}   
   \vspace{-1em}
   \caption{ \label{fig:model} Architecture of the proposed Model}
\end{figure*}

In addition to architectural hybrids, scale discrepancies caused by geographic variations in ground-sampling distance demand flexible multiscale feature extraction \cite{zhang2023efficient}. Transformers’ fixed input size and large parameter count also increase the risk of overfitting and inefficiency on small datasets \cite{zhang2023lightweight}. Recent strategies aim to address these challenges. RASNet leverages implicit neural representations (INRs) to model images as continuous functions, reducing sensitivity to resolution changes \cite{chen2022resolution}, while RSMamba employs a state-space model with hardware-aware design to achieve efficient global receptive fields at linear complexity \cite{chen2024rsmamba}. Although effective for sequence modeling, RSMamba lacks explicit multiscale convolutional mechanisms needed for spatial variability. Furthermore, many methods still rely on pretrained models or multisensor fusion, which increases computational cost and reduces adaptability \cite{roy2024cross}.

To address the above challenges, we propose a novel deep learning architecture for remote sensing scene classification that is based on the convolutional mixer paradigm. Unlike conventional CNNs or transformer based models, which separately emphasize local feature extraction or global dependency modeling, the proposed network uses convolutional mixer blocks to efficiently disentangle and process spatial and channel information. By combining channel mixing and spatial mixing convolutional operations, the model captures both detailed local structures and long range contextual patterns in remote sensing imagery, improving robustness against scale variations, background complexity, and limited labeled data while maintaining computational efficiency. Mixer networks, first introduced through the MLP-Mixer, separate spatial and channel processing using multilayer perceptrons, avoiding convolutions or attention while still modeling global dependencies \cite{tolstikhin2021mlp}. Their lightweight design has shown promise in aerial scene recognition, and PolSAR data analysis. For example, the MGCET model combines Mixers with graph convolution and transformers for improved robustness \cite{al2024mgcet}, and the PolSARConvMixer applies separate channel mixing and spatial mixing operations to capture polarization specific scattering mechanisms while reducing complexity \cite{jamali2024polsarconvmixer}. Building on these advances, our proposed framework aims to achieve a balance between accuracy, efficiency, and generalization, making it suitable for high resolution remote sensing scene classification across diverse datasets.

The main contributions of this work are as follows:  
\begin{itemize}
    \item Design of a convolutional mixer based architecture for remote sensing scene classification, which combines channel mixing and spatial mixing convolutional operations to jointly capture local structures and global context.  
   
    \item Extensive evaluation on multiple remote sensing datasets, demonstrating that the proposed model achieves a strong balance between accuracy, robustness, and efficiency compared with state-of-the-art methods.  
\end{itemize}
The remainder of this paper is organized as follows. Section \ref{sec:Methodology} presents the proposed model in detail. Section \ref{sec:results} outlines the experimental setup and discusses the results. Finally, Section \ref{sec:conclusion} concludes the paper by summarizing the key findings and highlighting future research directions.

\section{NETWORK ARCHITECTURE} 
\label{sec:Methodology}
The proposed model, illustrated in Fig.~\ref{fig:model}, begins with a patch embedding stage where the input image is divided into non-overlapping patches using a $4 \times 4$ convolution with stride 4, generating patch tokens of dimension $N=128$. To capture spatial dependencies at multiple scales, depthwise convolutional layers with kernel sizes $3 \times 3$ and $5 \times 5$ are applied. The resulting feature maps are then processed through a $1 \times 1$ convolution followed by GELU activation and batch normalization, which performs channel mixing and enhances the discriminative capacity of the learned representations. This process is repeated $L$ times, allowing the network to progressively refine spatial–channel interactions and build hierarchical representations of increasing complexity. A global average pooling (GAP) layer aggregates the spatial features into a compact descriptor, which is subsequently fed into a fully connected classification head with Softmax activation to produce the final predictions over $N$ categories.

\subsection{Depthwise Convolution}
\begin{figure}[!t]
\centering
\includegraphics[clip=true, trim = 20 10 20 5,width= 0.9\linewidth]{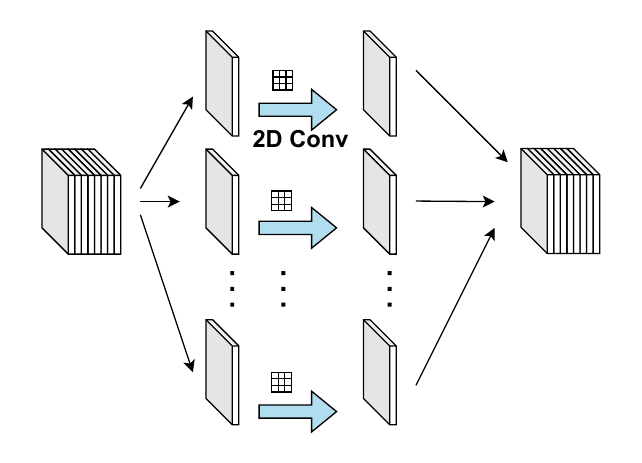}
\vspace{-1em}
\caption{Depthwise Separable Convolution: Input channels are separated, and each is convolved with a spatial filter. The split channels are then concatenated.}
\vspace{-1.25em}
 \label{fig:dwConv}
\end{figure}

In contrast to standard 2D convolution, which simultaneously combines information from all channels, depthwise convolution processes each channel with its own filter independently. This design significantly lowers the number of parameters and computational requirements while keeping the output channel dimension unchanged. By separating spatial and channel operations, depthwise convolution enables efficient extraction of spatial features within each channel, making it highly suitable for scenarios with limited computational resources. An illustration of this operation is provided in Fig.~\ref{fig:dwConv}.

\section{Experiments and Analysis}
\label{sec:results}
The proposed model is compared against several state-of-the-art deep learning based models, including 2D-CNN \cite{maggiori2016convolutional}, ViT \cite{dosovitskiy2020image}, ResNet-50 \cite{he2016deep}, MobileNet \cite{howard2017mobilenets}, EfficientNet \cite{tan2019efficientnet}, and VGG19 \cite{simonyan2014very}. All baseline models were trained and evaluated following their standard experimental setups as reported in the original publications to ensure fairness of comparison.
The model was trained for 100 epochs with a batch size of 32, and validation accuracy was monitored at the end of each epoch. To ensure optimal performance, the network was restored to the weights corresponding to the highest validation accuracy, with the best-performing weights automatically saved during training for reproducibility. The learning rate was initialized at $1 \times 10^{-3}$ and adaptively reduced by a factor of 0.5 whenever the validation accuracy failed to improve for 10 consecutive epochs, with a lower bound of $5 \times 10^{-5}$. Optimization was performed using the Adam optimizer. All models were implemented in Python using the Keras framework with TensorFlow as the backend, and to ensure a fair comparison, they were trained under identical conditions.

\subsection{Datasets}
\subsubsection{AID Dataset}
The AID dataset \cite{xia2017aid} is a large-scale aerial image benchmark constructed from Google Earth samples. Land use and land cover (LULC) maps derived from optical aerial photographs often share similar spatial structures and textures with the original imagery, since they are typically generated through visual interpretation. The dataset contains 10,000 images grouped into 30 scene categories, including beach, airport, baseball field, bare land, bridge, and church. Each image has a fixed size of $600 \times 600$ pixels, with spatial resolution ranging from 0.5 m to 8 m per pixel depending on the source imagery. Compared with single-source datasets such as UC-Merced, AID poses a greater classification challenge because its images originate from multiple remote sensing sources, introducing larger intra-class variation and inter-class similarity.

\subsubsection{EuroSAT}
The EuroSAT dataset \cite{helber2019eurosat} is a large-scale benchmark created using Sentinel-2 satellite imagery under the Copernicus program. It contains 27,000 labeled images organized into 10 land use and land cover (LULC) classes, including residential, industrial, highway, river, forest, and pasture. Each image has a fixed size of $64 \times 64$ pixels with a spatial resolution of 10 meters, covering 13 spectral bands that range from visible to shortwave infrared. The RGB images were generated by combining the red (B04), green (B03), and blue (B02) bands. A key strength of EuroSAT is its diversity, as the samples were collected from different regions across Europe, ensuring variability in climate, vegetation, and landscape conditions. These characteristics make EuroSAT particularly valuable for evaluating the generalization capability of classification models in remote sensing.

\begin{figure}[!t]
\centering
\includegraphics[clip=true, trim = 20 10 20 5,width= 0.9\linewidth]{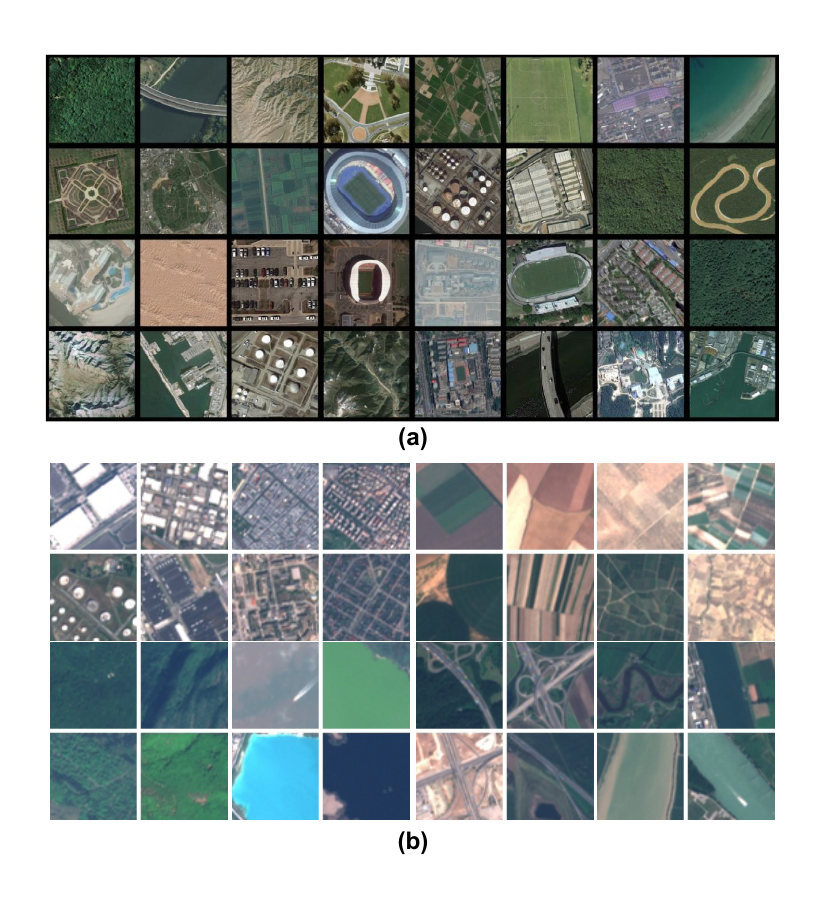}
\vspace{-1em}
\caption{Datasets used in this research, (a) Samples from AID dataset; (b) Samples from EuroSAT dataset}
\vspace{-1.25em}
 \label{fig:datasets}
\end{figure}

For each dataset, 70\% of images were randomly selected for training, 15\% for validation, and the remaining 15\% for testing, which is a common practice in this field. For the AID dataset, each image was resized to $64 \times 64$ pixels in order to reduce computational requirements. Table \ref{tab:dataset} summarizes the number of samples in each split, while Fig. \ref{fig:datasets} presents example images from different scene categories.

\begin{table}[!t]
\centering
\caption{Number of Training, Validation, and Testing Images for Each Dataset}
\vspace{-1em}
\label{tab:dataset}
\resizebox{0.95\linewidth}{!}{
\begin{tabular}{lcc}
\hline
\textbf{Dataset}                      & AID       & EuroSAT       \\ \hline
\textbf{Classes}                      & 30        & 10            \\
\textbf{Train Samples per Class}      & 154 - 294 & 1,400 - 2,100 \\
\textbf{Validation Samples per Class} & 33 - 63   & 300 - 450     \\
\textbf{Test Samples per Class}       & 33 - 63   & 300 - 450     \\
\textbf{Total Images}                 & 10,000    & 27,000        \\ 
\textbf{Spatial Resolution (m)}            & 0.5 - 8    & 10        \\ \hline
\end{tabular}
}
\end{table}

\subsection{Evaluation Metrics}
The classification performance of the developed models is evaluated using overall accuracy (OA), average accuracy (AA), and Cohen’s Kappa coefficient ($\kappa$). The overall accuracy measures the proportion of correctly classified samples among all test samples, and is defined as  
\begin{equation}
\text{OA} = \frac{\text{Number of correctly classified samples}}{\text{Total number of samples}} .
\end{equation}

The average accuracy provides the mean accuracy across all classes, ensuring that each class contributes equally regardless of its size. It is computed as  
\begin{equation}
\text{AA} = \frac{1}{n}\sum_{i=1}^{n} \frac{\text{True Positive}_{i} + \text{True Negative}_{i}}{\text{Total}_{i}} ,
\end{equation}
where $n$ is the number of classes.

To account for the possibility of agreement occurring by chance, Cohen’s Kappa coefficient is used. It is defined as  
\begin{equation}
\kappa = \frac{P_{o} - P_{e}}{1 - P_{e}} ,
\end{equation}
where $P_{o}$ denotes the observed agreement between predicted and true labels, and $P_{e}$ represents the expected agreement by chance.

\subsection{Classification Results}
The results in Table~\ref{tab:AID_results} demonstrate that the proposed model obtained the highest values across all evaluation metrics on the AID dataset, reaching an overall accuracy of 74.7\%, an average accuracy of 74.57\%, and a Kappa coefficient of 73.79. In comparison, VGG19 achieved 59.01\% overall accuracy, while ViT and MobileNet recorded 56.07\% and 55.34\%, respectively. ResNet50 and EfficientNet both remained below 51\% overall accuracy despite their larger size, suggesting that heavy networks with a high number of parameters may suffer from overfitting when trained on datasets with limited samples. On the EuroSAT dataset, the results in Table~\ref{tab:EuroSAT_results} indicate that the proposed model again achieved the best performance, reaching an average accuracy (AA) of 93.93\%, which outperformed all competing models. Among these, 2D-CNN and MobileNet showed strong results at 91.12\% and 90.10\%, respectively, while other architectures, including ResNet50, ViT, and VGG19, achieved moderate results in the range of 78--82\%, and EfficientNet recorded the lowest performance at 67.94\%.

It is also observed that the EuroSAT dataset generally produced higher accuracy scores across most deep learning models compared with the AID dataset. This can be attributed to the larger number of available images in EuroSAT, which helps improve generalization during training. In contrast, the AID dataset is collected from multiple sources with varying spatial resolutions, making it a more challenging benchmark for classification tasks. These differences explain why the overall performance on AID is lower and highlight the importance of dataset characteristics in influencing model performance.

\begin{table}[!t]
\centering
\caption{Classification Results Over the AID Data Benchmark}
\vspace{-1em}
\label{tab:AID_results}
\resizebox{0.95\linewidth}{!}{
\begin{tabular}{lccc}
\hline
\textbf{Model} & \textbf{OA (\%)} & \textbf{AA (\%)} & \textbf{$\kappa \times100$} \\ \hline
2D-CNN         & 70.56             & 69.97            & 70.62                       \\
ViT            & 56.07            & 56.24            & 54.50                        \\
ResNet50       & 50.40             & 50.16            & 48.63                       \\
MobileNet  & 55.34            & 54.80             & 53.75                       \\
EfficientNet & 48.13            & 47.81            & 46.28                       \\
VGG19        & 59.01            & 59.14            & 57.55                       \\
SceneMixer       & 74.70             & 74.57            & 73.79                       \\ \hline
\end{tabular}
}
\end{table}
\begin{table}[!t]
\centering
\caption{Classification Results Over the EuroSAT Data Benchmark}
\vspace{-1em}
\label{tab:EuroSAT_results}
\resizebox{0.95\linewidth}{!}{
\begin{tabular}{lccc}
\hline
\textbf{Model} & \textbf{OA (\%)} & \textbf{AA (\%)} & \textbf{$\kappa \times100$}\\ \hline
2D-CNN                      & 91.04 & 91.12 & 90.03 \\
ViT                         & 79.60  & 79.40  & 77.32 \\
ResNet50                    & 82.02 & 81.86 & 80.01 \\
MobileNet             & 90.07 & 90.10 &  88.96     \\
EfficientNet              & 68.72 & 67.94 & 65.26 \\
VGG19                     & 78.74 & 78.32 & 76.35 \\
SceneMixer                    & 93.90  & 93.93 & 93.22 \\ \hline
\end{tabular}
}
\end{table}

\begin{figure}[!t]
\centering
\includegraphics[width= 0.9\linewidth]{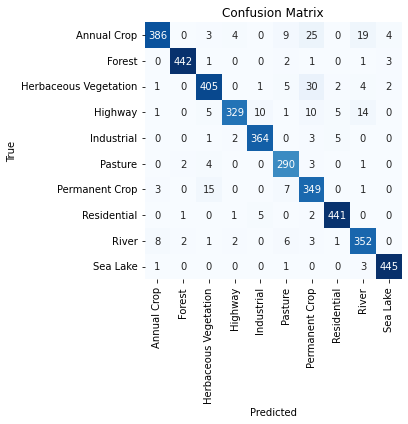}
\vspace{-1em}
\caption{Confusion Matrix of EuroSAT predictions}
\vspace{-1.25em}
 \label{fig:conf}
\end{figure}

The confusion matrix of the EuroSAT classification results obtained with the SceneMixer model is presented in Fig.~\ref{fig:conf}. It can be observed that most scene categories were classified correctly, with high percentages along the main diagonal. In particular, classes such as Forest, Residential, and Sea Lake show strong recognition, as indicated by the concentration of predictions in their respective diagonal cells. Some confusion is visible between visually similar categories, for example, Annual Crop and Permanent Crop, as well as between Highway and River, which share common features in the imagery. Overall, the matrix provides a detailed view of both the strengths of the model in separating distinct classes and the challenges that remain in distinguishing scenes with overlapping visual characteristics.

\subsection{Efficiency Analysis}
The results in Table~\ref{tab:efficiency} provide a comparison of different models in terms of memory usage and computational requirements. The number of parameters reflects the memory footprint of each model, while FLOPs (floating point operations) and MACs (multiply–accumulate operations) are directly related to computational cost. Larger models such as ResNet50 and VGG19 contain more than 23 million parameters and require over 634M FLOPs and 316M MACs, which indicates high memory usage and intensive computation. In contrast, lighter models such as MobileNet and EfficientNet reduce both parameters and operations, although their computational cost remains higher than the smallest models.  

The proposed model requires only 100,117 parameters while maintaining a FLOP count of 45.9M and a MAC count of 22.8M, values that are comparable to the much smaller 2D-CNN baseline. This shows that the proposed model substantially lowers memory requirements while keeping computational cost low, providing a more efficient alternative to conventional deep learning architectures.

\begin{table}[!t]
\centering
\caption{Parameters, FLOPs, and MACs of each Model used in the research}
\vspace{-1em}
\label{tab:efficiency}
\resizebox{0.95\linewidth}{!}{
\begin{tabular}{lccc}
\hline
\textbf{Model} & \textbf{Parameters} & \textbf{FLOPs} & \textbf{MACs} \\ \hline
2D-CNN         & 31,966              & 45,063,680     & 22,422,784    \\
ViT            & 1,006,046           & 60,697,088     & 18,158,592    \\
ResNet50       & 23,852,693          & 634,475,712    & 316,923,840   \\
MobileNet  & 2,425,822           & 49,342,648     & 24,657,372    \\
EfficientNet & 4,217,409           & 65,047,472     & 32,415,896    \\
VGG19        & 23,853,854          & 634,478,016    & 316,923,840   \\
SceneMixer       & 100,117             & 45,913,344     & 22,807,808    \\ \hline
\end{tabular}
}
\end{table}

\section{Conclusion}
\label{sec:conclusion}
The experimental results demonstrated that the proposed convolutional mixer model achieves strong performance for remote sensing scene classification while maintaining very low memory usage and computational cost. On the AID dataset, the model reached an overall accuracy of 74.7\%, an average accuracy of 74.57\%, and a Kappa value of 73.79, outperforming several widely used CNN and transformer baselines. On the EuroSAT dataset, the proposed model further improved performance, achieving 93.90\% overall accuracy, 93.93\% average accuracy, and a Kappa value of 93.22. These results confirm that combining spatial mixing through depthwise convolutions and channel mixing through pointwise operations provides an effective and efficient framework for scene classification across datasets with different characteristics.  

Despite these encouraging findings, several challenges remain for future research. The proposed approach does not yet incorporate attention mechanisms, which could enhance the ability to capture long-range dependencies and improve robustness to complex spatial patterns. In addition, more extensive comparisons against recent state-of-the-art architectures, including advanced hybrid CNN-transformer models, would provide a more comprehensive evaluation of the method’s competitiveness. Future work will also focus on testing the model against other challenging datasets such as NWPU-RESISC45 and UCMerced, as well as datasets that include hazy or dusty conditions such as RRSHID. Exploring these directions will help validate the model’s robustness and generalization ability in more diverse and realistic remote sensing scenarios.

\bibliographystyle{IEEEtran}
\bibliography{Main_Doc}

\end{document}